\documentclass{article}

\usepackage{microtype}
\usepackage{graphicx}
\usepackage{subfigure}
\usepackage{booktabs} 
\usepackage{subcaption}
\usepackage{url}

\usepackage{hyperref}

\newcommand{\E}{\mathbb{E}}
\newcommand{\indep}{\perp \!\!\! \perp}
\newcommand{\maximize}[2]{\underset{#1}{\text{maximize}} && #2}
\newcommand{\subjectto}[1]{\text{subject to} && #1}

\usepackage[accepted]{icml2024}

\usepackage{amsmath}
\usepackage{amssymb}
\usepackage{mathtools}
\usepackage{amsthm}

\usepackage[capitalize,noabbrev]{cleveref}

\theoremstyle{plain}

\theoremstyle{definition}

\theoremstyle{remark}

\usepackage[textsize=tiny]{todonotes}

\newcommand{\Ex}[1]{\E\left[#1\right]}
\icmltitlerunning{The Missing Link: Allocation Performance in Causal Machine Learning}

\begin{document}

\twocolumn[
\icmltitle{The Missing Link: Allocation Performance in Causal Machine Learning}



\icmlsetsymbol{equal}{*}

\begin{icmlauthorlist}
\icmlauthor{Unai Fischer-Abaigar}{lmu,mcml}
\icmlauthor{Christoph Kern}{lmu,mcml,mary}
\icmlauthor{Frauke Kreuter}{lmu,mcml,mary}
\end{icmlauthorlist}

\icmlaffiliation{lmu}{Department of Statistics, LMU Munich, Munich, Germany}
\icmlaffiliation{mcml}{Munich Center for Machine Learning, Munich, Germany}
\icmlaffiliation{mary}{Joint Program in Survey Methodology, University of Maryland, College Park, US}

\icmlcorrespondingauthor{Unai Fischer-Abaigar}{Unai.FischerAbaigar@stat.uni-muenchen.de}

\icmlkeywords{Automated Decision-Making, Distribution Shifts, Causal ML, CATE, Reliable Machine Learning}

\vskip 0.3in
]



\printAffiliationsAndNotice{}  

\begin{abstract}
Automated decision-making (ADM) systems are being deployed across a diverse range of critical problem areas such as social welfare and healthcare. Recent work highlights the importance of causal ML models in ADM systems, but implementing them in complex social environments poses significant challenges. Research on how these challenges impact the performance in specific downstream \emph{decision-making} tasks is limited. Addressing this gap, we make use of a comprehensive real-world dataset of jobseekers to illustrate how the performance of a single CATE model can vary significantly across different decision-making scenarios and highlight the differential influence of challenges such as distribution shifts on predictions and allocations.

\end{abstract}

\section{Introduction}
\label{introduction}

The usage of automated decision-making (ADM) systems is rising across a variety of critical problem domains \cite{fischerabaigar2024bridging}, including criminal justice \cite{angwin2016machine, mckayPredictingRiskCriminal2020}, child abuse prevention \cite{chouldechovaCaseStudyAlgorithmassisted2018a}, tax audit selection \cite{blackAlgorithmicFairnessVertical2022}, credit scoring \cite{kozodoiFairnessCreditScoring2022} and profiling of jobseekers \cite{desiereUsingArtificialIntelligence2021, kortnerPredictiveAlgorithmsDelivery2021}. Recent work has emphasized the importance of adopting causal machine learning (ML) methods in ADM systems over solely relying on predictive models which often prove inadequate for addressing counterfactual questions important for informed decision-making \cite{costonCounterfactualRiskAssessments2020b, feuerriegel2024causal, fernandez-loriaCausalDecisionMaking2022}. However, effectively implementing causal ML in decision-making systems presents a range of critical challenges \cite{fischerabaigar2024bridging}; among them dealing with diverse real-world objectives and constraints \cite{levyAlgorithmsDecisionMakingPublic2021, mitchellAlgorithmicFairnessChoices2021, passiProblemFormulationFairness2019}, mitigating distribution shifts between training and deployment environments, the difficulty of measuring accurate ground truth in complex decision scenarios \cite{guerdanGroundLessTruth2023a}, the need for uncertainty quantification \cite{bhattUncertaintyFormTransparency2021b} and explainability \cite{amarasingheExplainableMachineLearning2023a} as key factors to facilitate effective interaction between human decision-makers and ML systems \cite{enarssonApproachingHumanLoop2022}. 

While these challenges have in part been explored in the context of causal ML, there is limited research that focuses on evaluating the impact of these challenges on performance in specific downstream \emph{decision-making} tasks. To bridge this gap, we make use of a comprehensive real-world dataset containing high-quality information on a large cohort of jobseekers. This allows us to evaluate the performance of causal models across different decision contexts, and the differential influence of challenges such as distribution shifts on predictions and allocations.
We thereby highlight the unique intricacies of causal decision-making and the need for causal ML methodology tailored to the decision-making task.

\section{Causal Decision-Making}

In this paper, we focus on ADM systems used to solve resource allocation problems. In these settings, a decision-maker needs to distribute a limited social good or resource among a population of individuals. For example, a hospital might need to prioritize patients for a limited number of ICU beds, or an employment agency might need to identify which jobseekers require additional support. More specifically, we assume that a decision-maker needs to decide which individuals should receive a specific intervention. A common goal is to find the most efficient allocation of interventions with regards to an overarching utility. Causal machine learning models can aid in this task by estimating how the effectiveness of an intervention varies across different individuals. 

\subsection{Challenges in Connecting Decision-Making and Causal ML}

Deploying causal ML models to improve decision-making processes in complex and dynamic societal contexts comes with a variety of challenges. It can be difficult to accurately translate the intended goals of a decision-maker into a formalized ML system. Better estimation is not an end goal in itself, rather, the focus should be on how a model can best support the decision-making process. Recent work has identified key technical challenges where disjunctions between causal ML models and the goals of decision-makers tend to arise \cite{fischerabaigar2024bridging}. 

In this paper, we focus on the central challenge of distribution shifts between the training and deployment environment \cite{kouwIntroductionDomainAdaptation2018, gruberSourcesUncertaintyMachine2023a}. Such shifts can lead to a significant decline in model performance and result in unreliable outcomes, which is especially problematic in the dynamic environments where ADM systems are frequently deployed. Distribution shifts can arise from a variety of reasons \cite{moreno-torresUnifyingViewDataset2012}, including biased data collection processes, the deployment of a model in new geographic regions, or the continual deployment of the model itself \cite{perdomoPerformativePrediction2020}. Challenges at the intersection of causal ML and distribution shifts have been studied in prior technical studies \cite{johanssonGeneralizationBoundsRepresentation2022a, kuzmanovicEstimatingConditionalAverage2023b, kern2024multicate}. Unlike previous work that primarily focused on causal estimation, our aim is to emphasize the significance of evaluating challenges such as distribution shifts through the lens of the downstream decision-making task (i.e. causal decision-making).

\subsection{Problem Setting}

Following the potential outcomes framework, we denote a binary treatment $T_i \in \{0, 1\}$, covariates $X_i \in \mathcal{X}$ and two potential outcome $Y_i(0), Y_i(1) \in \mathbb{R}$. They describe the hypothetically possible outcome scenarios for each individual $i$: outcome under no intervention ($T_i = 0$), and the outcome if the intervention were allocated ($T_i = 1$). Given both potential outcomes, we could evaluate the individual treatment effect of an intervention as the difference $Y_i(1) - Y_i(0)$. However, for each individual it is only possible to observe the outcome in one reality, either $Y_i(0)$ or $Y_i(1)$, making it impossible to directly estimate the individual treatment effect. Given access to an observational dataset $\{(t_i, x_i, y_i)\}_{i=1}^n$, a common alternative is to estimate the so-called conditional average treatment effect (CATE), 
\begin{align}
    \tau(x) = \Ex{Y_i(1) - Y_i(0) | X_i = x}
\end{align}
for an individual with covariate vector $X_i = x$. Estimating the CATE function from observational data, requires a set of identifying assumptions: 1) \textit{Strong Ignorability} $\{Y_i(0), Y_i(1)\} \indep T_i | X_i$ 2) \textit{Positivity} $0 < P(T_i = 1 | X_i = x) < 1$ for all $x \in X_i$ 3) \textit{Consistency} $Y_i = Y_i(0)$, if $T_i=0$ and $Y_i = Y_i(1)$ if $T_i=1$. If these assumptions can be guaranteed, the CATE can be linked to a statistical estimand,
\begin{align}
\begin{split}
  \tau(x) = \Ex{Y_i | X_i=x, T_i=1} \\ - \Ex{Y_i | X_i=x, T_i=0}
\end{split}
\end{align}
allowing us to infer $\hat{\tau}(x)$ from data.

Given an estimated CATE model $\hat{\tau}(x)$, we consider three different decision-making scenarios. Within each scenario, our objective is to formulate an allocation policy $\pi: \mathcal{X} \to \{0, 1\}$, such that $\pi(x_i) = 1$ if the intervention is assigned to individual $i$ and $\pi(x_i) = 0$ otherwise. 

\begin{itemize}
    \item \textbf{Unconstrained Allocation (UC)} A decision-maker decides to intervene in all individual cases, for which the estimated CATE is positive $\hat{\tau}(x) > 0$. In such a scenario, no constraints on the number of interventions exist, making it only relevant to consider whether an intervention leads to a more beneficial outcome. For example, when designing an information campaign the cost of an individual message may be negligible, but it could still be relevant to estimate for which individual the messaging will have the desired effect \cite{athey2023machine}. Moreover, in a medical context, there may be a legal and moral imperative to aid all patients for which a treatment is expected to have a relevant net-benefit.
    \item \textbf{Top-K Allocation (TOP-K)} In the context of ADM, a common practice is to prioritize interventions for the top $k$ individuals who are ranked highest according to the CATE and for whom the effect is estimated to be positive. Formally, we assume a uniform cost for each intervention, with the allocation adhering to a budget constraint $\sum_{i=1}^{n} \pi(x_i) \leq k$. Such a constraint could represent practical limitations such as the availability of a fixed number of slots in a support program for jobseekers \cite{kernFairnessAlgorithmicProfiling2021}, or a hospital's limited capacity of ICU units.
    \item \textbf{Cost Efficient Allocation (CE)} In many scenarios, the cost of intervention can differ significantly depending on the individual case. For example, when the United States Internal Revenue Service (IRS) decides whom to audit, the complexity and resource requirements can vary strongly. For example, an in-person audit will be much more costly than a correspondence audit \cite{blackAlgorithmicFairnessVertical2022}. We consider a scenario where each possible intervention is associated with a different monetary cost $c_i \in \mathbb{R}$. The decision-maker must maximize the aggregated benefit while staying within the total available budget $C$. More formally, the optimal allocation policy is the solution of an integer program,
    \begin{align}
        &\maximize{\pi \in \{\mathcal{X} \to \{0, 1\}\}}{\sum_{i=1}^{n} \pi(x_i) \hat{\tau}(x_i)} \\
        &\subjectto{\sum_{i=1}^{n} \pi(x_i) c_i \leq C}
    \end{align}
    In our experiments, we use \texttt{optimize.milp} to solve this program, as implemented in the \texttt{SciPy} package.
\end{itemize}

\section{Methods}
\label{sec:methods}

We demonstrate the unique challenges of causal decision-making with an example situated in the labor market domain.
In recent years, public employment agencies have been increasingly deploying algorithmic profiling systems designed to allocate labor market support programs to jobseekers \cite{desiereUsingArtificialIntelligence2021, kortnerPredictiveAlgorithmsDelivery2021}. These systems often make use of predictive models trained on administrative data that is routinely collected by various governmental agencies.

\subsection{Dataset}

To conduct our empirical investigation, we obtained access to a large anonymized sample of German administrative labor market records \cite{siab2023, SchmuckerBerge2023SIABR}. This dataset, provided by the Research Data Centre (FDZ) of the German Federal Employment Agency (BA) at the Institute for Employment Research (IAB), contains a wide range of labor market activities from 1975 to 2021. The data includes individual information on employment history, benefit claims, periods of unemployment, and participation in job training programs for a large segment of the German population \cite{bachImpactModelingDecisions2023}.

For our analysis, we construct 13 covariates that include demographic (age, gender, citizen status and educational background) and employment-related features (such as cumulative days of unemployment and average salary over recent years). We selected a random sample of $5000$ jobseekers from the beginning of 2016 to train our (baseline) model. To evaluate the model's performance we use a separate sample of jobseekers from the beginning of 2018. For more details on the covariates and their distributions, refer to Appendix \ref{appendix}.

\subsection{Semi-Synthetic Simulations}

Since the effect outcomes $Y(1) - Y(0)$ can not be directly observed in real-world data, we make use of a semi-synthetic simulation setup inspired by \cite{curthNonparametricEstimationHeterogeneous2021, curth2023search}. We standardize our observed covariates $X$, and simulate outcomes for jobseekers using a linear model with higher-order interactions terms.

\begin{align}
    Y_i = c + \sum_j \beta_j X_j + \sum_{j, k} \beta_{jk} X_j X_k \\
    + \sum_{j, k, l} \beta_{jkl} X_j X_k X_l + T_i \sum_j \gamma_j X_j + \epsilon_i
\end{align}
where $\beta_j, \gamma_j \sim \text{Bern}(0.3)$, $\beta_{jk}, \beta_{jkl} \sim \mathcal{N}(0, 1)$ and $\epsilon_i \sim \mathcal{N}(0, 0.1)$. Following \cite{curth2023search}, each covariate is randomly part of one second order and third order interaction term. Treatment $T_i$ is allocated based on the propensity scores $\sigma(x) = 1 / (1 + e^{-\sum_j \beta_j X_j})$. Treatment represents a labor market support measure that needs to be allocated across a population of jobseekers with the aim of improving their employment prospects $Y_i$. 

For the third decision-making scenario (CE), we simulate individually varying costs using a log-normal distribution $\ln{c_i} \sim \mathcal{N}(-0.5, 1)$. By setting the mean to $1$, we facilitate easier comparison with the TOP-K setting. We assume that costs are known by the decision-maker. Alternatively, one may consider scenarios in which the expected costs need to be estimated by a separate prediction model.

\subsection{Distribution Shift}
\label{sec:shift}
Next to a baseline setting and a setting with limited training data, we evaluate the decision-making contexts under an external distribution shift between training data $\mathcal{D}_\mathcal{S}$ and test data $\mathcal{D}_\mathcal{T}$. This setup mimics real-world challenges encountered when implementing such systems, as labor market characteristics can change significantly between model deployment and training. Specifically, we introduce a covariate shift \cite{kouwIntroductionDomainAdaptation2018, moreno-torresUnifyingViewDataset2012}, similar to the approach described in \cite{kimUniversalAdaptabilityTargetindependent2022}. We estimate a propensity score model $\sigma(x)$ between training $\mathcal{D}_\mathcal{S}$ and test data $\mathcal{D}_\mathcal{T}$ using a random forest, as implemented in the \texttt{scikit-learn} package. We then strengthen the shift by an intensity of $q = 6$,
\begin{align}
    w(x) =  \text{logit}(\sigma(x))^q
\end{align}
and use the resulting weights $w(x)$ to sample a shifted training dataset. 

\subsection{CATE Estimation}

In recent years, a large variety of ML-based methods to estimate the CATE have been developed \cite{caronEstimatingIndividualTreatment2022, curth2023search}, with meta-learners \cite{kunzelMetalearnersEstimatingHeterogeneous2019} being particularly popular. In this context, we make use of the X-learner \cite{kunzelMetalearnersEstimatingHeterogeneous2019}, a method that is particularly well suited for dealing with unbalanced treatment groups. The X-learner first estimates two potential outcome functions $\mu_t(x) = \Ex{Y_i | T_i=t, X_i = x}$ for the treatment ($T_i = 1$) and control group ($T_i = 0$). Using these estimates, it constructs pseudo-outcomes for the treatment effect that are used to fit two separate CATE functions. Finally, the X-learner combines these CATE estimates by taking a weighted average, where the weights are determined by a propensity score model $\sigma(x) = \Ex{T_i = 1|X_i=x}$.

In our experiments, we use the X-learner implementation provided by the \texttt{EconML} package. As base learners, we use gradient boosting for the outcome regressions, and logistic regression for the propensity score model, as provided by the \texttt{scikit-learn} package. We use 5-fold cross validation to search over different hyperparameter values\footnote{Gradient Boosting: $\texttt{learning\_rate} \in \{0.1, 0.3\}$, $\texttt{max\_depth} \in \{3, 5, 8\}$ and $\texttt{n\_estimators} \in \{30, 100\}$. Logistic Regression: $\texttt{C} \in \{0.1, 1, 10, 100\}$. For all parameters that are not listed we use the default values of the \texttt{scikit-learn} implementation.}.

\section{Results}

\begin{figure}[ht]
\vskip 0.2in
\begin{center}
\centerline{\includegraphics[width=\columnwidth]{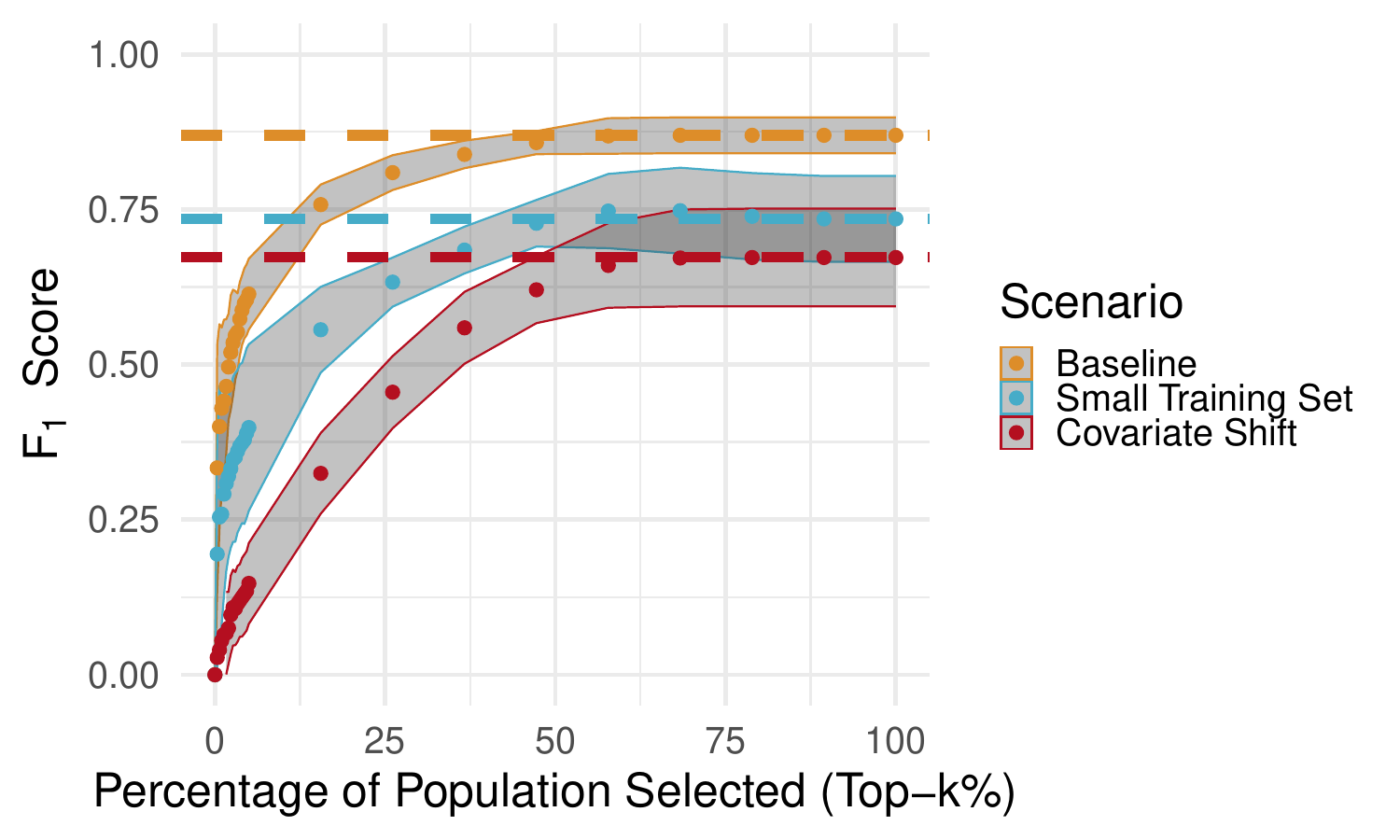}}
\caption{Comparison of Top-K Decision-Making Across Various Training Scenarios. The dashed lines correspond to the UC setting. Error ribbons indicate standard deviation over 10 outcome simulations.}
\label{f1-topk}
\end{center}
\vskip -0.2in
\end{figure}

\begin{figure}[ht]
\vskip 0.2in
\begin{center}
\centerline{\includegraphics[width=\columnwidth]{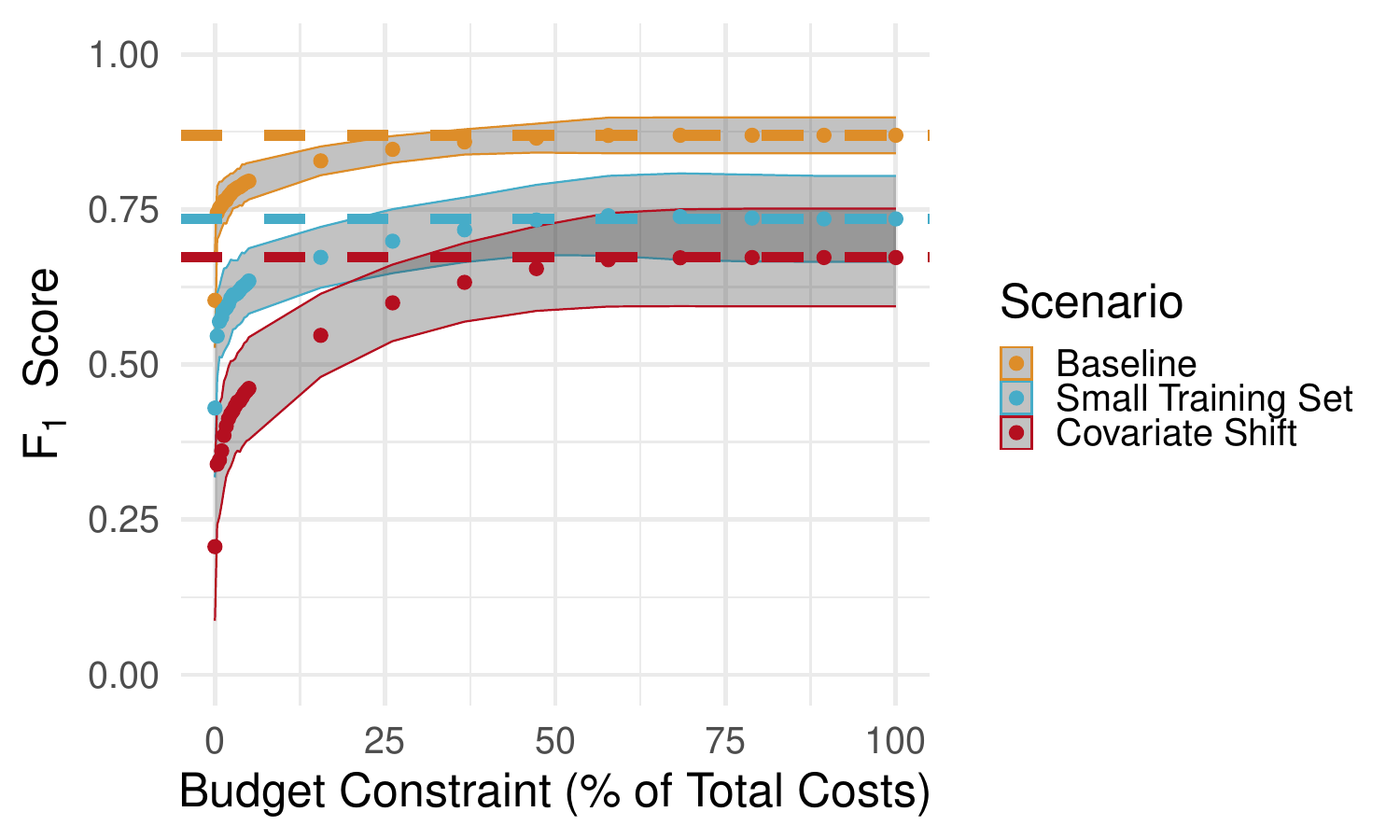}}
\caption{Comparison of CE Decision-Making Across Various Training Scenarios. The dashed lines correspond to the UC setting. Error ribbons indicate standard deviation over 10 outcome simulations.}
\label{f1-ilp}
\end{center}
\vskip -0.2in
\end{figure}

We evaluate the three decision-making scenarios (UC, TOP-K, CE) across three distinct settings: 1) A baseline setting with $n=5000$ training observations (Section \ref{sec:methods}), 2) A scenario with limited training data ($n = 500$) and 3) A scenario using a training dataset with a shifted covariate distribution, as described in Section \ref{sec:shift}. Our primary focus is on evaluating the quality of the downstream decision-making. For each scenario, we estimate the CATE on the test set, and determine for all individuals whether an intervention would be allocated. We then compare the resulting intervention group to the group that would have been selected if the decision-maker had access to the true treatment effects. The $F_1$ score is calculated to measure the allocation performance as the agreement between both groups.

Figures \ref{f1-topk} and \ref{f1-ilp} show the $F_1$ score under various budget constraints in the UC, TOP-K and CE setting, respectively. The accuracy of the decision-making considerably differs between settings, even when the budget constraints are comparable. For budget constraints $<50\%$, the UC decision-making is the most precise, which aligns with our expectations, since the procedure only requires accurate classification of the CATE. We observe that challenging training scenarios affect the decision-making contexts in different ways: a covariate shift results in a strong decline in allocation performance in the TOP-K setting for smaller budget constraints, whereas the UC decision-making (and to a lesser extent the CE setting) experiences a less significant impact. Note that such differences would not be identifiable by evaluating the predictive performance of the CATE model in isolation.

\section{Discussion}
In this study, we illustrate how the performance of a single CATE model can vary significantly across different decision-making scenarios using a combination of real-world data and synthetic outcomes. For example, while a distribution shift might threaten the ability of a model to accurately select the top 10\% of individuals, it could still provide fairly accurate classification of treatment effects. While we focus on a specific data-generating process and one external shift scenario, we can highlight how causal estimation and causal decision-making crucially differ.

Recent literature suggests that optimizing and evaluating (causal) models for decision-making in an end-to-end manner with regards to the decision-making objective results in better decision-making outcomes \cite{elmachtoubSmartPredictThen2022, fernandez-loriaCausalDecisionMaking2022}. We agree with this view and propose that future research should more explicitly focus on evaluating the reliability of CATE models in the context of the decision-making process they seek to inform. Ensuring the robustness of the output of the downstream decision problem under distribution shifts using tailored mitigation strategies may be more relevant in practice than solely aiming for predictive accuracy. Additionally, quantifying the uncertainty of the proposed decisions, not just predictions, could be a valuable direction for future research. This would involve exploring how model uncertainty propagates through various decision optimization tasks, particular in complex settings involving multiple objectives.


\section*{Acknowledgements}
This work is supported by the DAAD programme Konrad Zuse Schools of Excellence in Artificial Intelligence, sponsored by the Federal Ministry of Education and Research and by the Baden-Wuerttemberg Foundation under the grant “Fairness in Automated Decision-Making - FairADM”.


\bibliography{bibliography}
\bibliographystyle{icml2024}

\newpage
\appendix
\onecolumn

\section{Appendix}
\label{appendix}

\begin{table}[htbp]
\centering
\caption{List of Covariates}
\vspace{0.2in}
\label{tab:covariates}
\begin{tabular}{llp{10cm}}
\toprule
\textbf{Covariate Name} & \textbf{Type} & \textbf{Description} \\
\midrule
\textit{yearofbirth}            & Numerical & Year of birth. \\
\textit{age}         & Numerical & Age of the individual. \\
\textit{cumul\_ue}         & Numerical & Total number of days unemployed, while registered as a job seeker over the past four years. \\
\textit{cumul\_employed} & Numerical & Total number of days employed subject to social security over the past four years. \\
\textit{cumul\_marginal} & Numerical & Total number of days in marginal employment over the past four years. \\
\textit{cumul\_meas} & Numerical & Total number of days participating in individual support measures over the past four years. \\
\textit{cumul\_benefit} & Numerical & Total number of days receiving unemployment benefit over the past four years. \\
\textit{cumul\_train} & Numerical & Total number of days as trainee or intern over the past four years. \\
\textit{income} & Numerical & Daily wage or benefit. \\
\textit{gender} & Categorical & Gender.\\
\textit{germancitizen} & Categorical & German citizen status. \\
\textit{voc\_education} & Categorical & Vocational training qualification. \\
\textit{school} & Categorical & School leaving qualification. \\
\bottomrule
\end{tabular}
\end{table}

\begin{figure}[ht]
\vskip 0.2in
\begin{center}
\centerline{\includegraphics[width=\textwidth]{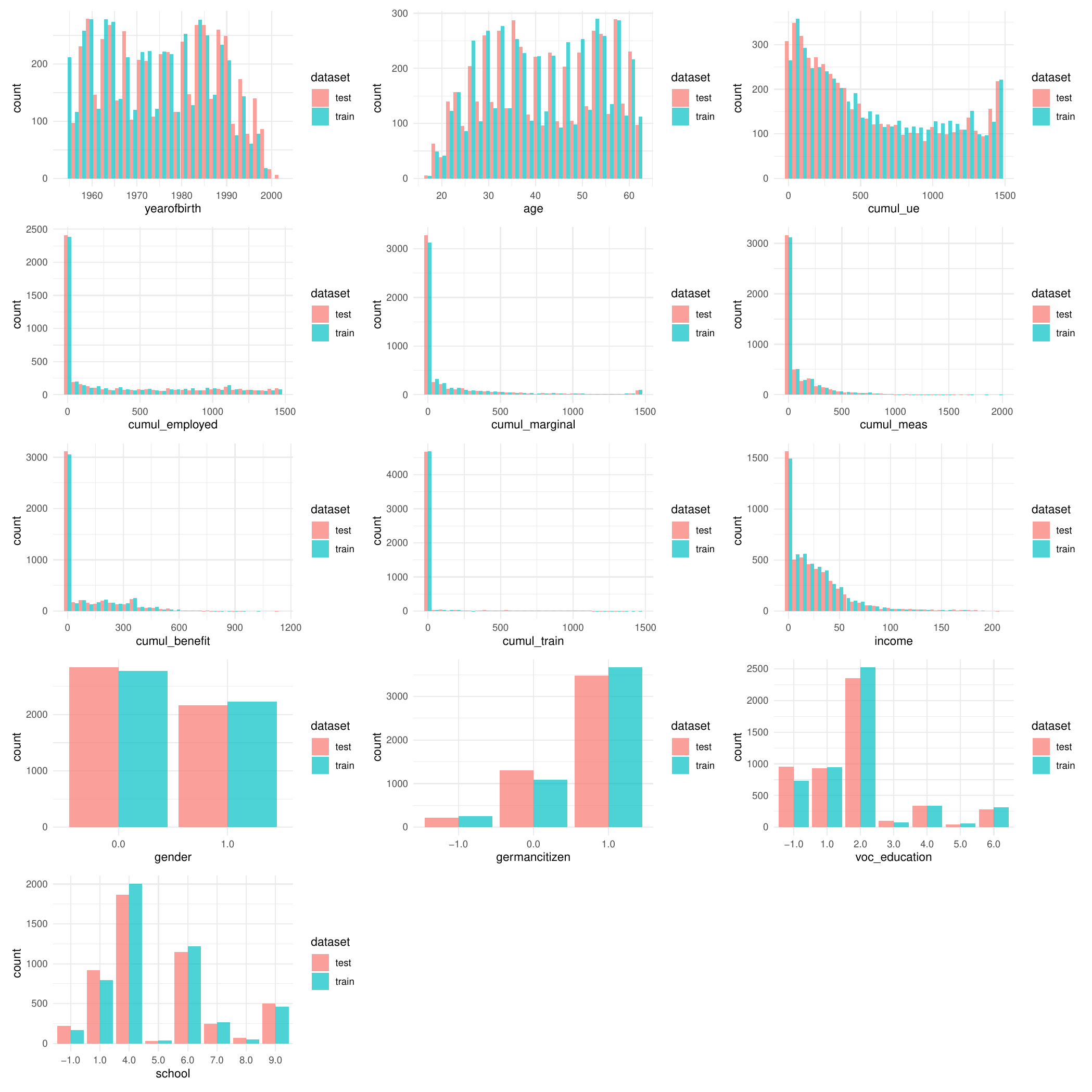}}
\caption{Covariate Distribution of Training and Test Dataset in the \emph{Baseline} Scenario.}
\end{center}
\vskip -0.2in
\end{figure}

\begin{figure}[ht]
\vskip 0.2in
\begin{center}
\centerline{\includegraphics[width=\textwidth]{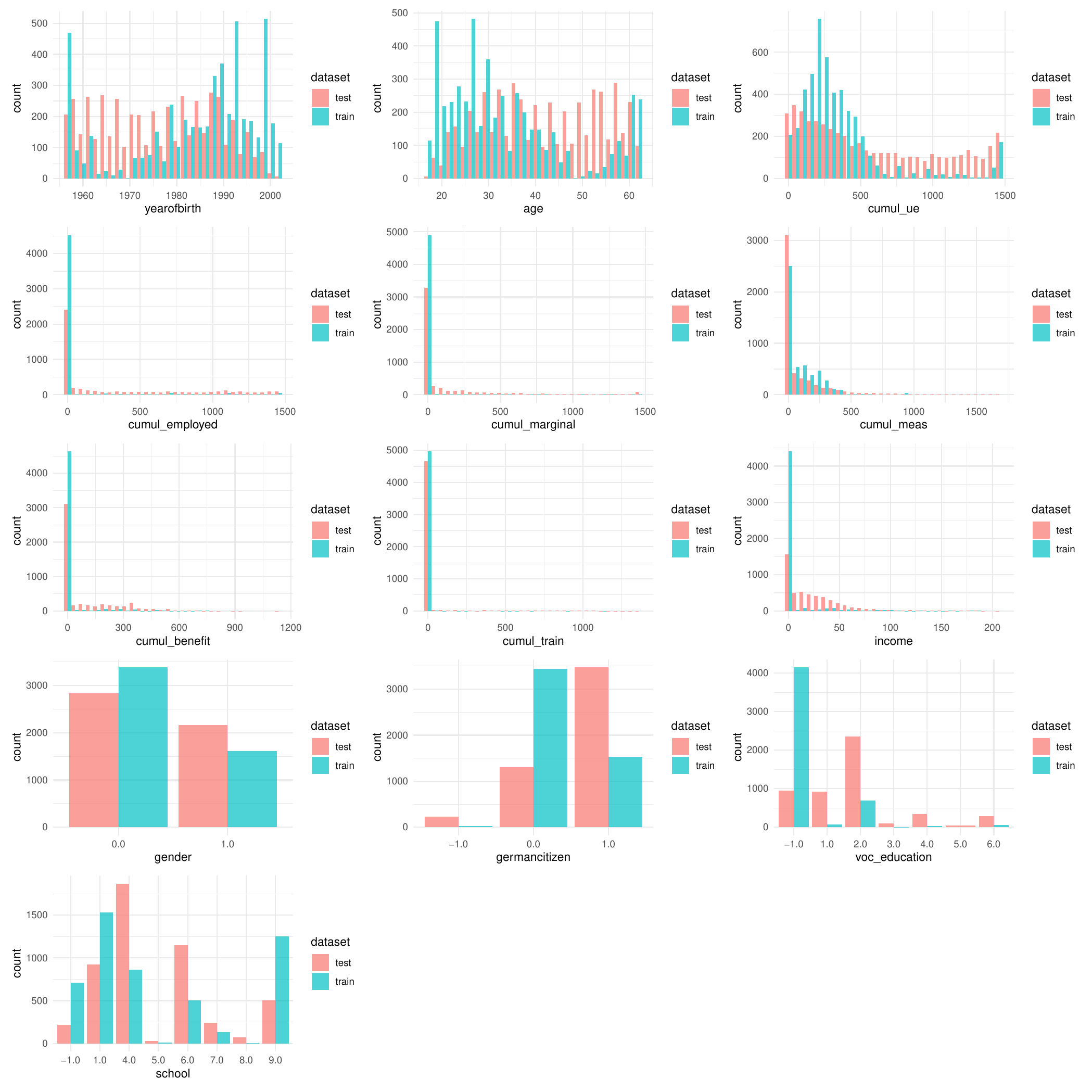}}
\caption{Covariate Distribution of Training and Test Dataset in the \emph{Covariate Shift} Scenario.}
\end{center}
\vskip -0.2in
\end{figure}



\end{document}